\documentclass[12pt]{IEEEtran}
\usepackage[utf8]{inputenc}
\usepackage{url}
\usepackage{cite}
\usepackage{amsmath,scalerel}
\usepackage{graphicx}

\title{Representation, Analysis of Bayesian Refinement Approximation Network: A Survey}

\author{
\IEEEauthorblockN{Ningbo Zhu\IEEEauthorrefmark{1}, Fei Yang\IEEEauthorrefmark{2}}\\
\IEEEauthorblockA{\IEEEauthorrefmark{1}University of Alberta. \emph{ningbo@ualberta.ca}}
\IEEEauthorblockA{\IEEEauthorrefmark{2}University of Alberta. \emph{fei5@ualberta.ca}}
}

\date{February 2021}

\begin{document}

\maketitle

\begin{abstract}

After an artificial model background subtraction, the pixels have been labelled as foreground and background. Previous approaches to secondary processing the output for denoising usually use traditional methods such as the Bayesian refinement method. In this paper, we focus on using a modified U-Net model to approximate the result of the Bayesian refinement method and improve the result. In our modified U-Net model, the result of background subtraction from other models will be combined with the source image as input for learning the statistical distribution.  Thus, the losing information caused by the background subtraction model can be restored from the source image. Moreover, since the part of the input image is already the output of the other background subtraction model, the feature extraction should be convenient, it only needs to change the labels of the noise pixels. Compare with traditional methods, using deep learning methods superiority in keeping details.

\end{abstract}


\section{Introduction}
With the utilization of deep learning networks, more and more different computer vision areas have recently achieved excellent performance on their research. Especially in background subtraction, the accuracy of the result is dramatically increasing in several years. Researchers use different deep learning methods to label every pixel in each frame of video and use traditional methods to secondary processing the output, hope to get a better result. But the traditional methods such as the Bayesian Refinement method, usually have a disadvantage when processing the output, which is losing details. Therefore, a general solution to secondary processing background subtraction output still remains a challenging problem that keeping result details.
\par
Essentially, secondary processing background subtraction's result is a process to denoise. Typically both the input image and the ground-truth image are binary masks. Newly machine learning-based denoise methods rarely handle the binary mask and the traditional denoise method based on pixel spatial relationship such as the Bayesian Refinement method can handle it but will lose image corner feature details. To solve this problem, we focusing on combining the machine learning method and the traditional method to design a Bayesian Refinement approximation network for pixel label Inference. This model has the function of the Bayesian Refinement method but can keep the corner feature details in the image.
\par
To build our model, we use a U-Net as a template. U-Net has the best performance for image segmentation which is widely accepted. When the U-Net is downsampling the input binary mask image, we are downsampling the source image which is not been background subtraction at the same time. Because the binary mask input image is generated by the other machine learning network using the source image, the binary mask image and the source image have the same image shape. Every time when the U-Net is downsampling, we do the same process to the source image, then combine it with the downsampling result and replace the downsampling result with our modified result. Therefore we restored the missing information from the source image which is caused by downsampling.

\section{Brief Summary of Existing Work}
This section introduces three state-of-the-arts related to our topic. Dynamic Deep Pixel Distribution Learning (D-DPDL) proposed by Zhao et al. in 2019 \cite{zhao2019dynamic} focuses on handling the noise in the binary mask after background subtraction. Dense U-net Based on Patch-Based Learning proposed by Wang et al. in 2019 \cite{wang2019dense} and Global-Feature Encoding U-Net (GEU-Net) proposed by Xiao et al. in 2020 \cite{xiao2020global} modified the original U-Net to supplement the information lost when the background is subtracted from the traditional model. The researches of these two papers are image segmentation and image fusion, but both of their models are based on U-Net, which can be used for reference.

\subsection{D-DPDL}

\begin{figure}[h!]
\centering
\includegraphics[scale=0.2]{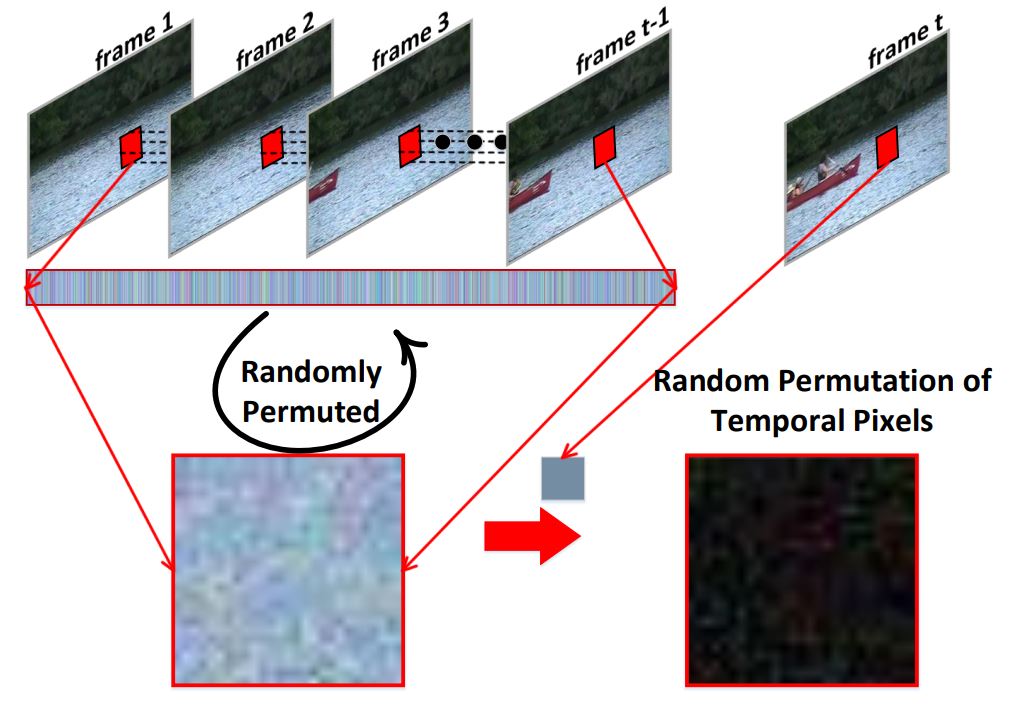}
\caption{The processing of extract RPoTP feature \cite{zhao2019dynamic}}
\label{fig:rpotp}
\end{figure}

The pixel of the same position in different frames will be permutated randomly, and then reshaped into a matrix. The Random Permutation of Temporal Pixels (RPoTP) feature is the difference between the matrix and the current status of that pixel. The processing is shown in Fig.\ref{fig:rpotp}. 

\begin{figure}[h!]
\centering
\includegraphics[scale=0.2]{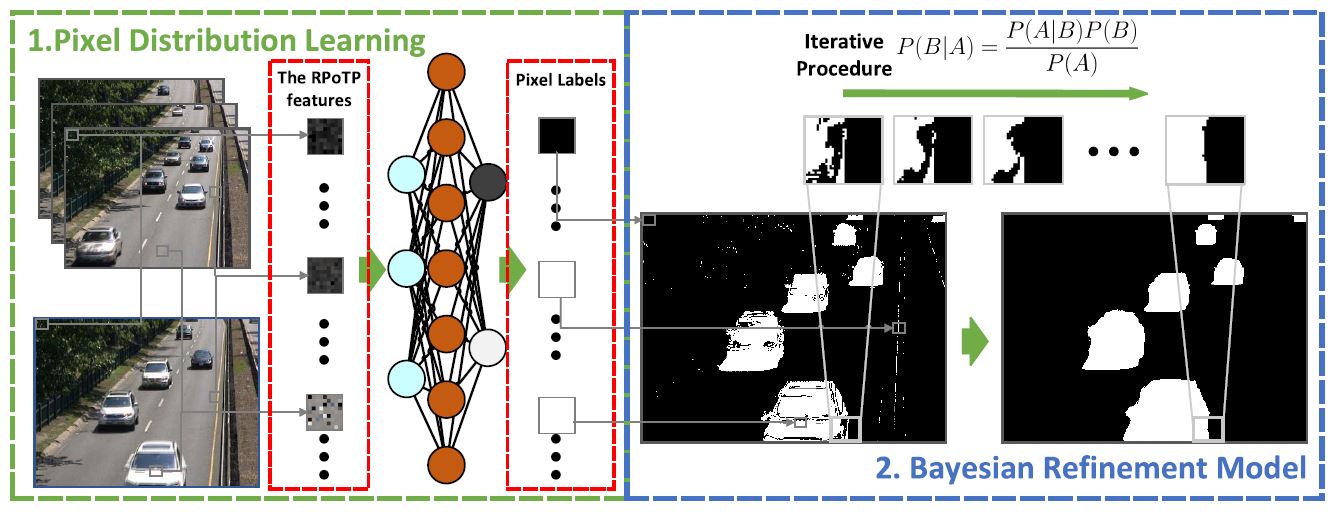}
\caption{The proposed approach of D-DPDL \cite{zhao2019dynamic}}
\label{fig:ddmodel}
\end{figure}

The D-DPDL model generates RPoTP features dynamically, then feed these features into a Convolutional Neural Network (CNN) train for a classifier for background subtraction.  There are some random noises in the outputs of the network due to the RPoTP features is introduced by randomly permutating, therefore the Bayesian Theorem is used to refine this noise in the result. The proposed approach is shown in Fig.\ref{fig:ddmodel}. Complex video images include the flow of light and shadow, intricate movement of objects, shadows, etc., which will cause a lot of inevitable noise when background subtraction. Bayesian refinement helps the D-DPDL model reduce these noises by binarization and the output result looks very close to the groundtruth. However, some detailed features at the edge of the foreground and background are over-smoothed so that some important information lost.

\subsection{Dense U-net Based on Patch-Based Learning}
The dataset is preprocessing before inputted into Dense U-Net. Patches extraction can help to solve the error-prone manually segment retinal vessel and save time. Each training image can be extracted N patches randomly and the patches extracted from testing images are used to calculate the average final segmentation result. The model can learn these features efficiently. Dense U-Net is composed of traditional U-Net added five dense blocks. Dense blocks help each layer take advantage of the input features. The result of Dense U-Net is better than U-Net, but there is still a big gap with the groundtruth, and there is still room for improvement. This method can obtain a better segmentation effect than state-of-the-arts and has clinical application value.

\subsection{GEU-Net}
GEU-Net is proposed to handle multi-focus image fusion problems, which cannot be solved by the traditional Convolutional Neural Network (CNN)-based multi-focus image fusion methods. The focusing ability of imaging equipment limits the acquisition of fully-focused images. Focusing only on local features is not enough to generate an accurate focus map. The global information across objects and local areas needs to be added to the feature information and used to restore edge information in the output. To adequately extract and apply global semantics and edge information, the authors proposed the global feature pyramid extraction module (GFPE) and global attention connection upsample module (GACU).

\section{Other Publish Review}

\subsection{Enhancing Videoconferencing Using Spatially Varying Sensing \cite{Enhancing}}
\begin{figure}[h!]
\centering
\includegraphics[scale=0.4]{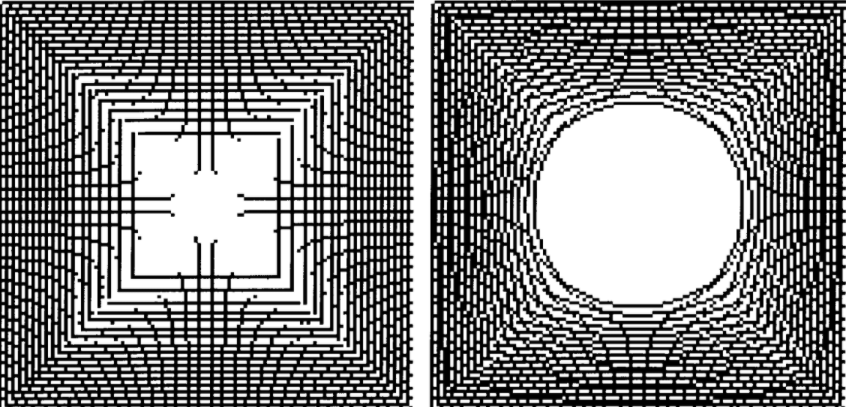}
\label{fig:Fovea}
\caption{Fixed distortion and conformable fovea versus variable distortion and circular fovea. A circular fovea is shown at right as opposed to a conformable fovea on the left.}
\end{figure}

In this paper, Dr.Basu introduces a new method to compress images. The method based on human vision can be characterized as a variable resolution system. The human eye can see the region around the fovea (point of attention) clearly, but the periphery is viewed
in lesser detail. In many images, there will be one or more areas that are more important than the rest of the picture. In such an area (fovea) more detail is needed. The outer regions
(periphery) are not that important, and thus less detail is needed. This decrease in detail within the periphery may be achieved by variable resolution (VR). \ref{fig:Fovea} 
VR compressed may lose some information in the secondary importance area, but it is still useful in some areas, such as videophone, and image database.

\begin{figure}[h!]
\centering
\includegraphics[scale=0.4]{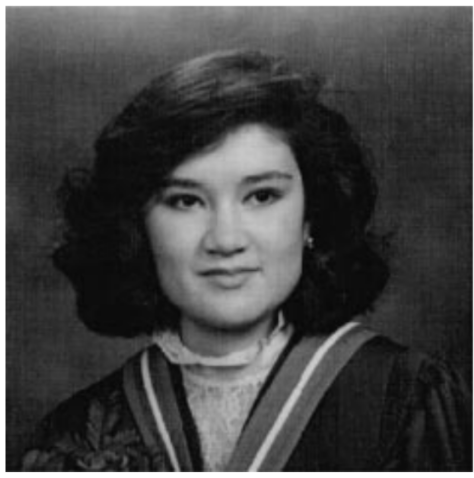}
\label{fig:Fovea_ori}
\caption{Original image}
\end{figure}

\begin{figure}[h!]
\centering
\includegraphics[scale=0.4]{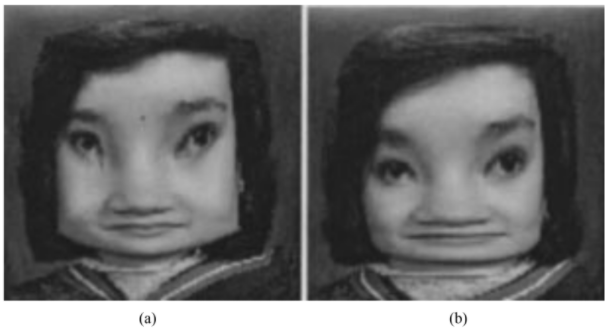}
\label{fig:Fovea_compressed}
\caption{Compressed image, fovea near center. compression is 90\%. (a) Modified and (b) Cartesian. }
\end{figure}

\begin{figure}[h!]
\centering
\includegraphics[scale=0.4]{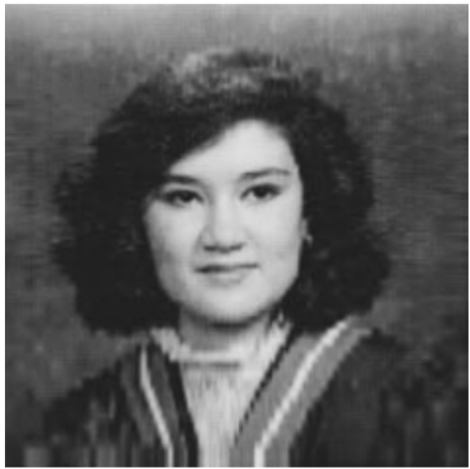}
\label{fig:Fovea_uncompressed}
\caption{Uncompressed image (CVR method). }
\end{figure}

\subsection{Hough transform for feature detection in panoramic images \cite{HoughTransform}}
Omni-directional sensors are useful in obtaining a 360 field-of-view. Not like a 360-degree camera which combines two 180-degree camera image, Omni-directional sensors can achieve this by using one single camera. The image obtained by Omni-directional sensors is 360-degrees, and it is not fit to use the traditional feature detection method on it. In this paper, Dr.Fiala and Dr.Basu introduce a new method Hough transform for feature detection in panoramic images. Hough transforms images from image space to Hough space, which points look like curve lines in it. Then a histogram of edge pixels by angle allows detection of radial lines, which are projections of lines parallel to the main camera axis. At last re-projection of corresponding points back to image plane loci for horizontal lines found from Panoramic Hough space, and vertical lines from peaks in the histogram.

\begin{figure}[h!]
\centering
\includegraphics[scale=0.4]{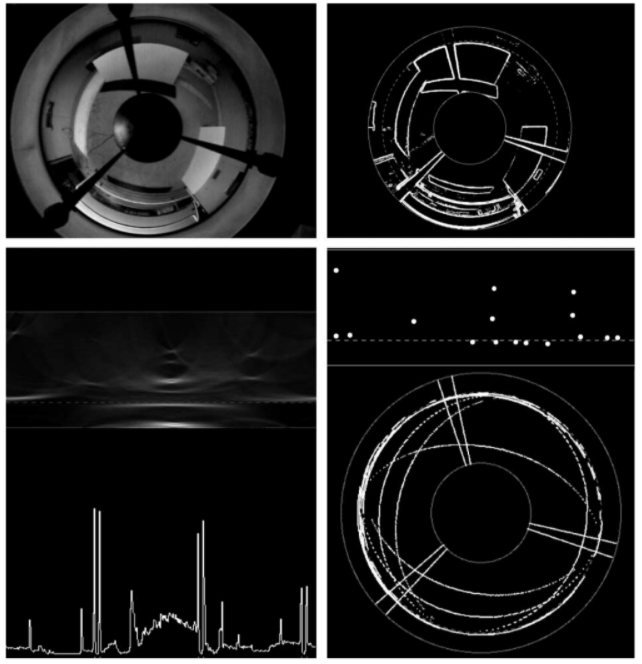}
\label{fig:HoughTrans}
\caption{Single lobe experiment (left to right, top to bottom): original image, edge magnitude image, panoramic horizontal and vertical Hough transform, automatic cluster detection of horizontal lines, re-projection of lines back onto image space for automatically detected lines. }
\end{figure}

\subsection{Perceptually Optimized 3-D Transmission Over Wireless Networks \cite{Perceptually}}
\begin{figure}[h!]
\centering
\includegraphics[scale=0.4]{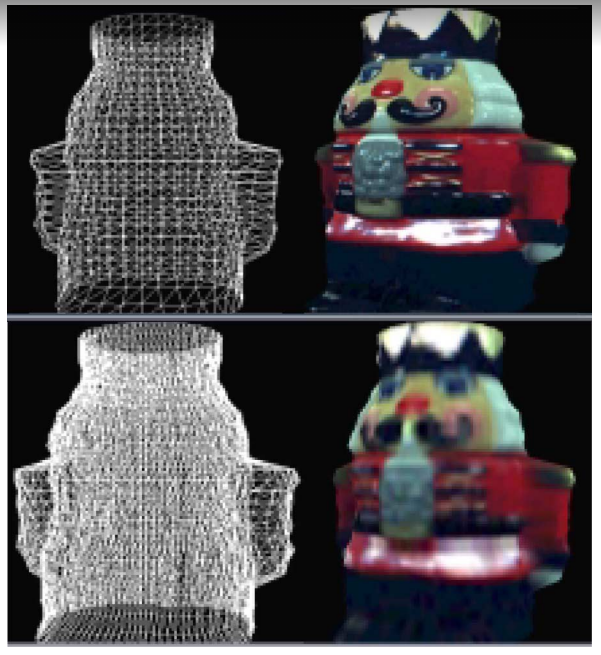}
\label{fig:3-DTransmission}
\caption{Top has lower quality mesh, requiring 125 Kb total bandwidth, and higher perceptual quality; Bottom has higher quality mesh, and lower quality texture requiring a total bandwidth of 134 Kb, but has lower perceptual quality. }
\end{figure}

Nowadays the size of high-quality 3-D models is quite big. The 3-D models maybe damaged hardly by wireless transmission. In this paper, Dr.Cheng and Dr.Basu introduce a theoretical framework for determining the relative importance of texture versus mesh was presented. They outlined an approach for estimating perceptual quality considering variations in mesh and texture resolutions. In this theory, only one of the mesh and texture needs to be high quality, the quality of the final model can be good enough. It can reduce the size of the 3-D model and increase the robustness for wireless transmission at the same time. \ref{fig:3-DTransmission}

\subsection{NOSE SHAPE ESTIMATION AND TRACKING FOR MODEL-BASED CODING \cite{Nose}}
Feature extraction on the face plays an important role in applications of model-based coding and human face recognition. In the old way, eyes and mouth are considered to be the most important features contributing to different facial expressions. However, detecting and tracking the nose shape plays an equally important role as eyes and mouth for model-based coding, especially for analysis and synthesis of realistic facial expressions. In this paper, Dr.Yin and Dr.Basu proposed a nose shape detection method for facial recognition. The important nose features lie in the shapes of nostrils and nose sides. The accurate shape of the nostril and the nose side can be important for composing a facial expression, especially when a person is smiling or laughing, the nostril shape and the nose side are obviously changed. The nostril has distinctive darkness from the facial skin, which colour-based region growing can roughly detect the position and the approximate shape of the nostrils. The nose side has a shape like a vertical parabola. After the nostril detection, the nose side can be detected easily with the initial comer of the nose side(using information from nostrils).

\subsection{Perceptually Guided Fast Compression of 3D Motion Capture Data \cite{PercetuallyGuided}}
In this paper Dr.Firouzmanesh, Dr.Cheng, and Dr.Basu introduced a fast encoding and decoding technique for lossy compression of motion capture data based on human perception. Experimental results show that the algorithm in this paper is much faster than other comparable methods. The study suggests that the compression ratios of at least 25:1 are achievable with little impact on perceptual quality. The method in this paper is faster while preserving an equivalent or better perceptual quality, the compressed motion data is more robust to constrained bandwidth, which is important in a mobile environment. In 3D Motion Capture, there can be many other factors affecting the quality of animations, including the distance from the camera, horizontal and vertical velocity of the object and the size of the limbs. In the model provided by the paper, Dr.Firouzmanesh, Dr.Cheng, and Dr.Basu introduce two attention factors: bone lengths and variation in rotation. 

\subsection{Integrating active face tracking with model-based coding \cite{Intergrating}}
In this paper, Dr.Yin, and Dr.Basu introduce a new method of real-time face tracking. In this method, input from an active camera is used for MPEG4 model-based coding. At first, considering a moving camera, the background is compensated. Then by using frame differences fusion, the talking face is segmented from the compensated background.  After that, a morphological filter is then applied to make the system less sensitive to noise. Without this step, the facial features may be detected wrongly. Then, using the Hough Transform and deformable template coupled with colour information to detect the facial features. At last, an adapted wireframe model is prepared for the extracted face. 

\subsection{Enhancement of low visibility aerial images using histogram truncation
and an explicit Retinex representation for balancing contrast and color consistency \cite{liu2017enhancement}}
This paper presents a new way to ariel images under low visibility which is by improving multi-scale Retinex (MSR) based enhancement.  This paper extends the multi-scale Retinex with more than three scales. An explicit multi-scale representation based on mathematical analysis and deductions can balance image contrast and colour consistency. Then this paper introduces a new technique that can post-processing strategy to remap the multi-scale Retinex output to the dynamic range of the display.

\subsection{Event Dynamics Based Temporal Registration \cite{singh2007event}}
In this paper, Basu et al. (2007) argue that use event dynamics can generate high temporal resolution video for both global and local temporal registration of video sequences. event dynamics is a property that is inherent to an event and is thus common to all acquisitions of the event. Basu et al. proved that a global event dynamics-based approach can have smaller temporal registration errors.

\subsection{Noise2Noise \cite{N2N}}
Based on DIP \cite{DIP}, N2N introduced a theorem to train a CNN model learning image restoration without clean data. From this paper, there is a point of view when training a network for image restoration and denoise, the ground truth does not need to be 100\% clean data. If we have several noised images which by adding noise to one single ground truth image. The network can still get similar restoration results when training with a noise-noise image pair(the same image has different noise). In another word, the average of the different noised signals is close to the ground truth signal. In this method, the limit is, if the noise-noise image pair loss too much information from ground truth, the model cannot restore those information. The model doesn't gather information from outside, it only gathers information from the noise images. It cannot create information for the restoration.

\subsection{Noise2Void \cite{N2V}}
Based on Noise2noise, noise2void introduces a new network structure “blind-spot network”. Traditionally if we train such a network using the same noisy image as input and as target, the network will degenerate and simply learn the identity. In a blind-spot network, the receptive field of each pixel excludes the pixel itself, preventing it from learning the identity. Blind-spot network doesn’t need noise-noise image pairs to train, it can learn to remove pixel-wise independent noise when they are trained on the same noisy images as input and target. Noise2Void has the same problem as Noise2Noise, it cannot gather information outside of the image and restore it. And also, Noise2Void using a mask with a hole in the center to achieve this network structure. Because the blind-spot network uses the surrounding pixels to conjecture the center pixel, it will make the restoration lose some special information, such as sparkles. And the output will look a little bit blur.

\subsection{Noise2Self \cite{N2S}}
Continuing the idea from Noise2void, Noise2self introduces an idea to calculate the loss for model training. From the experiment, the Noise2self team found out, if we use a radius=3 donut shape median filter to both noise image and ground truth image, the loss of two images between before and after is similar. So if we have only one noisy image, then we can use this radius=3 donut shape median filter to blur it and calculate the loss. Then use this loss to backpropagation and train our model to get a similar result.

\subsection{A Multisensor Technique for Gesture Recognition Through Intelligent Skeletal Pose Analysis \cite{rossol2015multisensor}}
 In this paper, a novel multisensor technique that improves the pose estimation accuracy during real-time computer vision gesture recognition has been introduced. The method using a premeasured artificial hand to trained an offline classifier and learn which hand positions and orientations are likely to be associated with higher pose estimation error. By using the artificial hand, this new method can reduce total pose estimation error by over 30\%.

\section{Conclusion}
The interest in background subtraction has dramatically increased in the last few years. Though traditional methods still dominate in this area, machine-learning-based algorithms appear as one of the most attractive solutions to go beyond the limits. In this survey, we have presented a comprehensive view of the solutions for the representation, analysis and recognition of machine-learning-based background subtraction. Though this area has attracted considerable interest-only quite recently, which is continuously growing. More and more literature has been published in this area, it gives us more research directions.

\bibliographystyle{IEEEtran}
\bibliography{references}

\end{document}